\documentclass{article}

\usepackage{PRIMEarxiv}

\usepackage[utf8]{inputenc} 
\usepackage[T1]{fontenc}    
\usepackage{hyperref}       
\usepackage{url}            
\usepackage{booktabs}       
\usepackage{amsfonts}       
\usepackage{nicefrac}       
\usepackage{microtype}      
\usepackage{lipsum}
\usepackage{fancyhdr}       
\usepackage{graphicx}       
\graphicspath{{media/}}     
\usepackage[T1]{fontenc}    
\usepackage{hyperref}       
\usepackage{url}            
\usepackage{picins}
\usepackage{wrapfig}
\usepackage{booktabs}       
\usepackage{amsfonts,bm}       
\usepackage{nicefrac}       
\usepackage{microtype}      
\usepackage{xcolor}         
\usepackage{ulem}
\usepackage[inline, shortlabels]{enumitem}
\usepackage{todonotes}
\usepackage{changes}
\usepackage{float,subcaption}
\usepackage{caption}
\usepackage{todonotes}
\usepackage{makecell}
\usepackage{tabulary}
\usepackage{tabularx}
\usepackage{stackengine}
\usepackage{adjustbox}
\usepackage{diagbox}
\usepackage{tabstackengine}
\usepackage[T1]{fontenc}
\usepackage{amsmath,bm}
\usepackage{multirow}
\definechangesauthor[color=brown, name={Milad Leyli-abadi}]{ML}

\title{ML4PhySim : Machine Learning for Physical Simulations Challenge (The airfoil design)

}

\author{
     M. Yagoubi, M. Leyli-Abadi, D. Danan, J. Brunet \\
     IRT SystemX, Palaiseau, France
      \And  JA Mazari $^1$, F. Bonnet $^{1,2}$ \\ \small{1}. Extrality, Paris, France \\ \small{2}. Sorbonne Université, CNRS, ISIR \\ \And A. Farjallah \\ NVIDIA 
     \And M. Schoenauer \\ INRIA, Saclay, France \And P. Gallinari \\ Sorbonne Université, CNRS, ISIR \\ Criteo AI Lab
}
\newcommand*{\scorecircle}[2]{
    \begin{tikzpicture}[scale=0.15]%
    \draw (0,0) circle (1);
    \fill[fill opacity=0.5,fill=#2] (0,0) -- (90:1) arc (90:90-#1*3.6:1) -- cycle;
    \end{tikzpicture}
}

\newcommand*{\myfullcircle}[1]{
    \begin{tikzpicture}[scale=0.1, line width=.3mm]]%
    \draw[black, fill=#1] (0,0) circle (1.5);
    \end{tikzpicture}
}

\begin{document}
\maketitle

\begin{abstract}
The use of machine learning (ML) techniques to solve complex physical problems has been considered recently as a promising approach. However, the evaluation of such learned physical models remains an important issue for industrial use. The aim of this competition is to encourage the development of new ML techniques to solve physical problems using a unified evaluation framework proposed recently, called Learning Industrial Physical Simulations (LIPS). We propose learning a task representing a well-known physical use case: the airfoil design simulation, using a dataset called AirfRANS. The global score calculated for each submitted solution is based on three main categories of criteria covering different aspects, namely: ML-related, Out-Of-Distribution, and physical compliance criteria. To the best of our knowledge, this is the first competition addressing the use of ML-based surrogate approaches to improve the trade-off computational cost/accuracy of physical simulation.The competition is hosted by the Codabench platform with online training and evaluation of all submitted solutions \footnote{\href{https://www.codabench.org/competitions/1534/}{https://www.codabench.org/competitions/1534/ }}.
\end{abstract}


\section{Competition description}

\subsection{Introduction}
Nowadays, numerical simulation represents an essential tool in designing and managing physical complex systems, thanks to its lower cost compared to  real-world experimentations. In most cases, classical numerical approaches  can predict the physical behavior of the systems accurately. However, the corresponding  computational cost is often very high, thus prohibiting their use in complex industrial contexts. The ML approaches have been  successfully used to solve a broad range of problems such as computer vision, natural language processing, and voice recognition. Recently,  Deep Learning (DL) approaches have seen  a growing interest in their application on various physical domains where the numerical analysis approaches are hard to design or involve costly and imprecise computations. These approaches have gained popularity due to their ability to solve complex tasks, leading to   promising results in various physical domains (see e.g.,\cite{tompson2016,kasim2021building,menier2023cd,rasp_deep_2018,sanchez2020learning, liu2021multi}). They allow also an important speed-up of simulations by substituting some computational bricks with data-driven numerical models.
The aim of the proposed challenge is to encourage the development of new ML solutions to solve physical problems. To do so, 
 we propose to learn a well-known CFD use case : \textit{the airfoil design}.
Our objective is to allow an efficient benchmarking and evaluation of the submitted solutions. For this purpose, we rely on our recently proposed  benchmarking platform called LIPS (Learning Industrial Physical Simulation) \cite{leyli2022lips}.

The goals of this challenge are to:
\begin{itemize}[nolistsep,leftmargin=*]
    \item Promote the use of ML  models to solve physical problems, by exploring several hybridization strategies. 
    
    \item  Promote  transparency and industrial relevance by integrating  a dedicated  reproducibility approach of ML for physics:  participants will be required to submit untrained models, which will then be fully retrained and evaluated on our servers. This will ensure that the models are reproducible and can be tested independently, thereby enhancing the transparency of the entire process;
    \item Encourage collaboration between AI and physical sciences by designing new ML approaches (algorithms, architectures) targeting  physical problems;
    \item Ensure a  \textit{meaningful} evaluation of the submitted ML surrogate models thanks to a homogeneous comparison of different physical tasks using the same multi-criteria evaluation proposed in the LIPS Framework \cite{leyli2022lips};
    \item Benchmark all submissions on the same environment, especially for \textit{fair} speed-up comparisons; 
     \item Foster collaboration and knowledge sharing between participants from different communities (ML and Physical science), through open discussions and feedback sessions:
    \item Ultimately, contribute to the development of more efficient, reliable, and cost-effective solutions for real-life physical systems, benefiting both industry and society as a whole.
\end{itemize}

\subsection{LIPS Framework}
\begin{figure}
    \centering
    \includegraphics[width=\linewidth]{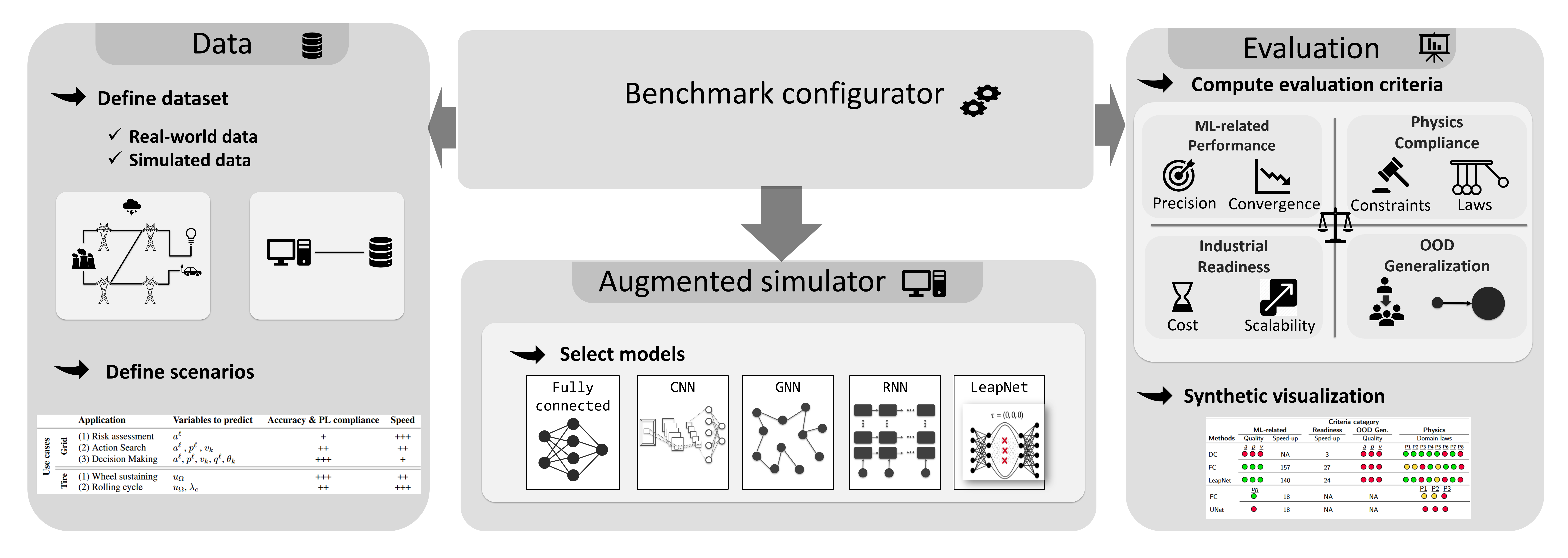}
    \caption{LIPS fraework}
    \label{fig:lips}
\end{figure}

LIPS \cite{leyli2022lips} is a unified extensible platform enabling the definition and evaluation of new physical use cases in homogeneous and yet flexible form. 
It consists of four modules combining data management, benchmark configurator, augmented simulator and evaluation (figure \ref{fig:lips}). 

The main objective of using LIPS in the ML4PhySim Challenge is the design of generic and  comprehensive  evaluation criteria to rank submitted solutions. This evaluation has to consider several aspects to represent industrial needs and expectations. Classical ML-related metrics are not sufficient in that regard. Thus,  three categories of criteria have been considered in this challenge. \\

\paragraph{ML-related performance} Among classical ML metrics, we focus on the trade-offs of typical model
accuracy metrics such as Mean Absolute Error (MAE) vs computation time. \\
\paragraph{Application-based out-of-distribution (O0D)} Generalization For industrial physical simulation,
there is always some expectation to extrapolate over minimal variations of the problem geometry/physical parameters
depending on the application. We hence consider OOD geometry evaluation such as unseen airfoil mesh variations. \\
\paragraph{Physics compliance} Physical laws compliance is decisive when simulation results are used to make
consistent real-world decisions. Depending on the expected level of criticality of the benchmark,
this criterion aims at determining the type and number of physical laws that should be satisfied.
Note that the  aforementioned airfoil use case is already implemented in this platform. A new use case and the benchmark are instantiated by selecting a dataset, an augmented simulator, and an evaluation object, as shown in Figure \ref{fig:lips}. Each module could be parameterized through a generic configuration file to make it more user-friendly. They further comply with simple interfaces, making it modular to add new evaluation metrics or new physical domains. This framework is hence the first pillar enabling the setup of our competition.

\subsection{Modulus}
NVIDIA Modulus \cite{modulus} is the second important framework we leverage in our competition. NVIDIA Modulus is an open-source, freely available, AI framework for building, training, and fine-tuning Physics-ML models with a simple Python interface. One can build high-fidelity AI surrogate models that blend the causality of physics described by governing partial differential equations (PDEs) with simulation data from CAE solvers or observed data. Such AI models can predict with near-real-time latency and for a parameterized design space.


\section{Physical simulation scenario: the airfoil design}

The competition will address the challenge of improving baseline solutions for the airfoil design problem by building ML-based surrogate models. The overall aim is to reduce the physical simulation cost while preserving acceptable precision for the outputs.  Furthermore, we encourage solutions that can be generalized to solve other scenarios of the airfoils usecase throught the OOD generalization dataset. 


%
\begin{figure}[htb]
  \begin{center}
    \begin{subfigure}{.45\textwidth}
      \centering
      \includegraphics[width=\linewidth]{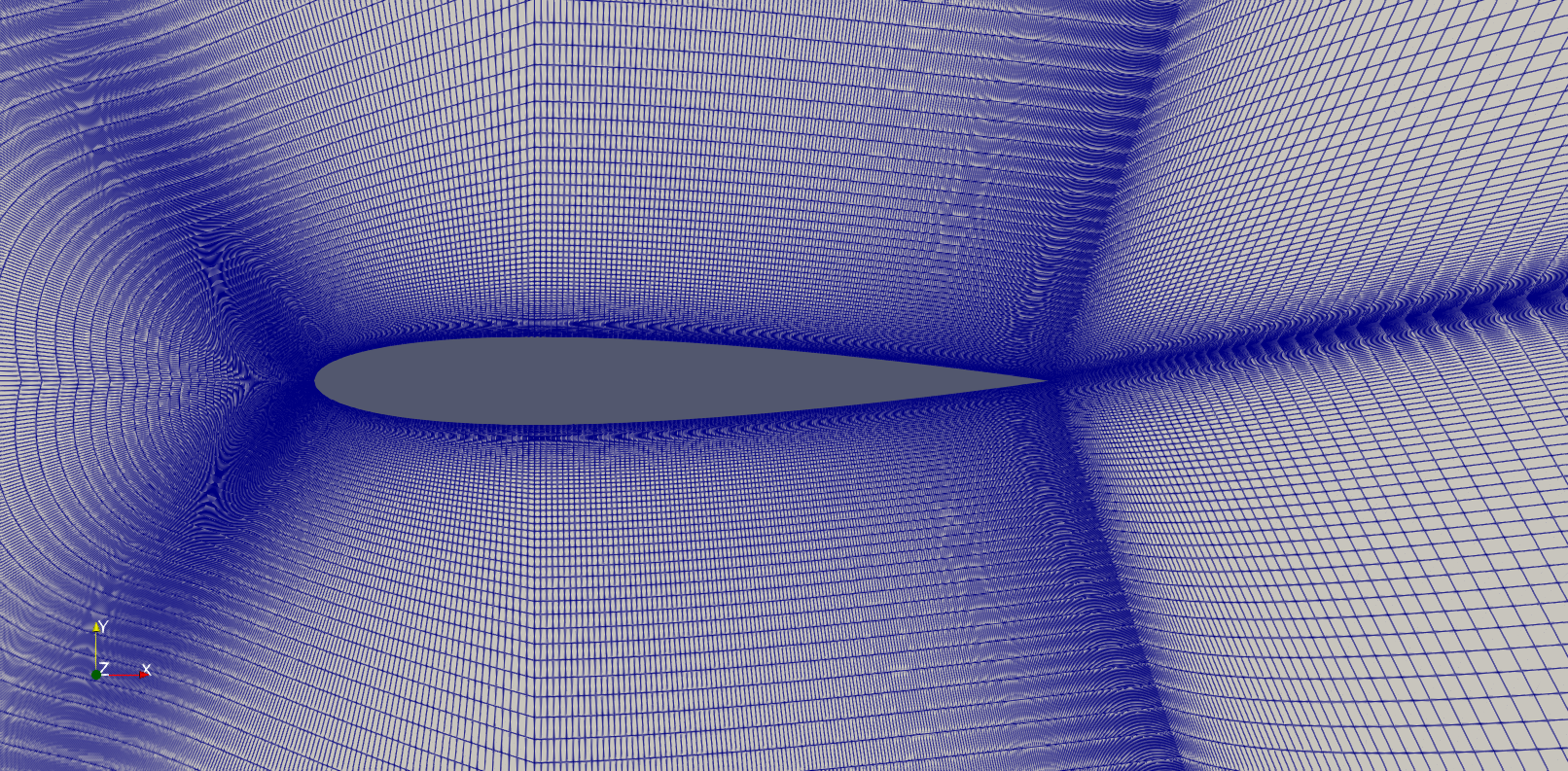} 
      \caption{Input}
    \end{subfigure}
    \begin{subfigure}{.45\textwidth}
      \centering
      \includegraphics[width=\linewidth]{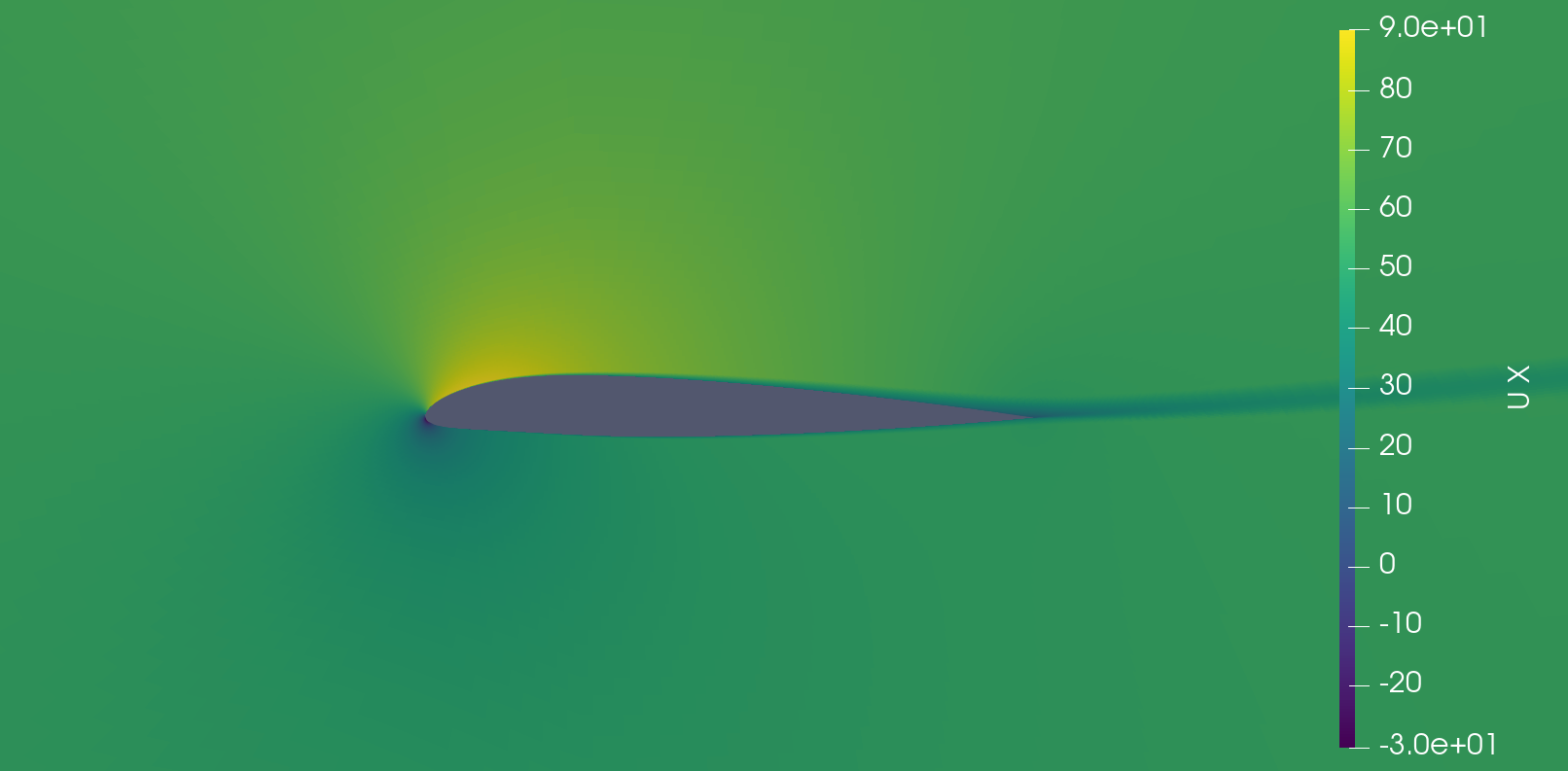}
      \caption{Output}
    \end{subfigure}
    \end{center}

    \caption{(a) Airfoil mesh. (b) $x$-velocity field. 
    }
    \label{fig:cfd_gdl}

\end{figure}

\subsection{Problem description}
The conception and design of planes require a rigorous study of the surrounding physical fields that could be measured experimentally. However, it is time-consuming and computationally expensive to set up prototypes which could be dangerous. Moreover, prototypes don't allow the reproduction of highly complex configurations. To circumvent that, one of the solutions is virtual testing which enables coping with complex constraints and testing configurations that could not be possible in reality.
Hence, numerical simulation is crucial for modeling such physical phenomena that are governed by \textit{Partial Differential Equations} (PDEs) and widely used in \textit{Computational Fluid Dynamics} (CFD) to solve \textit{Navier-Stokes equations} (NS). NS-PDEs are highly nonlinear and their analytical resolution is out of reach.  They are numerically solved with the help of discretization methods such as finite differences, finite elements, or finite volumes methods. 
At high Reynolds number (beyond a certain threshold), NS-PDEs involve a complex dissipation process that cascades from large length scales to small ones (\textit{Kolmogorov} microscales) which makes their direct resolutions challenging and expensive (thousands of CPU hours). Following this complexity and the related challenges, traditional CFD solvers apply numerical methods (and most of them rely on turbulence modeling) to solve these equations including \textit{Direct Numerical Simulation} (DNS) which is prohibitively expensive since no turbulence model is used, \textit{Large Eddy Simulation} (LES) to model the smaller scales of turbulence and \textit{Reynolds-Averaged Navier-Stokes} (RANS) that solve mean field equations and allow to model all scales of turbulence.

 Despite the efficiency of these methods, NS-PDEs are still computationally expensive and may take several hours to converge to an accurate solution w.r.t the granularity of meshes. For that reason, we argue that \textit{Deep Learning} (DL) could be a potential surrogate model to cover several CFD tasks including design exploration, design optimization, inverse problem, and real-time control, as well as super-resolution. DL enjoys several advantages: (i) can represent a large family of functions, (ii) can lead to mesh-free models, (iii) trade-off accuracy and complexity, and (iiii) fast inference once the model is trained.
 Regarding the aforementioned challenges and the potential of surrogate models, a benchmarking \textit{aerodynamic} dataset called \textit{AirfRANS} \cite{bonnet2022airfrans} is introduced to study the capabilities of DL in approximating the functional space of NS-PDEs (\textit{RANS} in this case along with k-$\omega$ SST model for turbulence modeling). This dataset is inspired by real-world phenomena and industrial use cases, and its design choice is validated relying on the experimental data produced by \textit{NASA} and available on the \textit{Turbulence Modeling Resource} (TMR) of the \textit{Langley Research Center} \cite{TMR}. The simulations are run with \textit{OpenFoam} \cite{Jasak07openfoam:a,OpenFOAM}.
From a design standpoint, the challenges are: (1) simulations come in the form of unstructured mesh with millions of nodes, (2) High Reynolds leading to sharp signals, (3) Difficulty at encoding the geometry and boundary conditions w.r.t complex topological and physical variations including Angle of Attacks (AOA).
In this challenge, we propose to study the AirFoil design problem by considering a scarce data regime \cite{bonnet2022airfrans}. The task consists in predicting the incompressible steady-state two-dimensional fields and the force acting over airfoils in a subsonic regime. The goal is to find the airfoil that maximizes the lift-over-drag ratio and predict the velocity and pressure fields around it accurately. 
To \textit{physically} evaluate the DL models, only surface and volume fields are regressed.
Then, 
force coefficients are computed 
as post-treatments to stick with the form of \textit{RANS} equations. Therefore, the trained DL model is said to be \textit{physically consistent} only  if the predicted fields and the derived quantities predictions are consistent. Figure \ref{fig:cfd_gdl} illustrates an example of the input/outputs of CFD solvers and DL models.
\subsection{Data}
As mentionned above, the AirfRANS dataset generated for the purpose of the competition is based on specific physical solver that represents the ground truth: OpenFoam (see table \ref{tab:input_output}). For more details about the datasets, the reader could refer to \cite{bonnet2022airfrans}.
For the needs of the challenge, the datasets will be slightly adapted, without any major changes. Each simulation is given as a point cloud defined via the nodes of the simulation mesh, that is to say a discretization of the 2D domain considered.
Inputs and outputs variables for each task are listed in the table \ref{tab:input_output}.
For the challenge, we consider three datasets each endowed with several samples to be used in the ML task, such as training and testing: 

\begin{itemize}
    \item \textbf{Training set}: 103 samples, AirfRANS 'scarce' task, training split, filtered to keep the simulation where the number of reynolds is between 3e6 and 5e6
    \item \textbf{Test set} : 200 samples, AirfRANS 'full' task, testing split
    \item \textbf{OOD test Set} : 496 samples, AirfRANS reynolds task, testing split
\end{itemize}
Note that each sample is associated to a CFD 2D simulation using OpenFoam.

\begin{table}[H]
    \centering
   \caption{Reference data for the task and its related input/output variables.}
    \resizebox{\columnwidth}{!}{
    \begin{tabular}{lllll}
         \toprule
         \textbf{Task} & \textbf{Reference physical simulator} & \textbf{Dataset description} & \textbf{Input variables} & \textbf{Output variables}  \\
         \midrule

         \multirow{2}{*}{Airfoil design} & \multirow{2}{*}{OpenFoam \cite{OpenFOAM}} & AirfRANS  \cite{bonnet2022airfrans} & Positions  & Velocity ($\bar{u}_x$, $\bar{u}_y$)  \\
            & & Documentation\footnote{\url{https://airfrans.readthedocs.io/en/latest/index.html}}& Inlet velocity & Pressure divided by the specific mass ($\bar{p_s}$)\\
               & & GitHub\footnote{\url{https://github.com/Extrality/AirfRANS}}& Distance to the airfoil & Turbulent kinematic viscosity ($\bar{\nu}_t$)\\
               & & & Normals  &  \\
         \bottomrule
    \end{tabular}}
    \label{tab:input_output}
\end{table}

\section{Metrics for solutions ranking}
We propose an \textit{homogeneous evaluation} of the submitted solutions to learn  the airfoil design task  using the LIPS (Learning Industrial Physical Systems Benchmark suite) platform proposed in \cite{leyli2022lips}. The evaluation is performed through 3 categories that cover several aspects of augmented physical simulations namely: 
\begin{itemize}
    \item ML-related: standard ML metrics (e.g. MAE, RMSE, etc.) and speed-up with respect to the reference solution computational time;
    \item Physical compliance: comparison between the physical criteria (post-processing involving the solver output) computed and the reference physical criteria
    \item Application-based context: out-of-distribution (OOD) generalization to  extrapolate over minimal variations of the problem  depending on the application; speed-up;

\end{itemize}
In the ideal case, one would expect a solution to perform equally well in all categories but there is no guarantee of that. In particular, even though a solution may perform well in standard machine-learning related evaluation, it is required to assess whether the solution also properly respects the underlying physics.
For each category, specific criteria related to the airfoils design task are defined. 
The global score is calculated based on linear combination  formula of the three evaluation criteria categories scores mentioned in equation \ref{eq: score}

\begin{equation}
\label{eq: score}
Score=\alpha_{ML}\times \bm{Score_{\textbf{ML}}} 
+ \alpha_{OOD}\times \bm{Score_{\textbf{OOD}}} + \alpha_{PH}\times \bm{Score_{\textbf{Physics}}},
\end{equation}
where $\alpha_{ML}$, $\alpha_{OOD}$ and $\alpha_{PH}$ are the coefficients to calibrate the relative importance of ML-Related, Application-based OOD, and Physics Compliance categories respectively.
We explain in the following subsections how to calculate each of the three sub-scores for each category. 

\subsection{"ML-related" score calculation}
This sub-score is calculated based on a linear combination form 2 sub-criteria namely: Accuracy and speedup.

\begin{equation}
\label{eq: scoreML}
Score_{ML}=\alpha_A\times \bm{Score_{\textbf{Accuracy}}}
+ \alpha_S\times \bm{Score_{\textbf{Speed}}},
\end{equation}
where $\alpha_A$ and $\alpha_S$ are the coefficients to calibrate the relative importance of accuracy and speedup respectively.

For each quantity of interest, the accuracy sub-score is calculated based on two thresholds that are calibrated to indicate if the metric evaluated on the given quantity gives unacceptable/acceptable/great result. It corresponds to a score of  0 point / 1 point / 2 points, respectively. Within the sub-cateogry, Let
\begin{itemize}
\item $Nr$, the number of unacceptable results overall 
\item $No$, the number of acceptable results overall
\item $Ng$, the number of great results overall
\end{itemize}
Let also $N$, given by $N=Nr+No+Ng$. The score expression is given by
\begin{equation}
\label{eq: score_ML_accuracy}
Score_{\textbf{Accuracy}}= \frac{1}{2N} (2\times Ng+1\times No +0\times Nr)
\end{equation}

A perfect score is obtained if all the given quantities provides great results. Indeed, we would have $N=Ng$ and $Nr=No=0$ which implies $Score_{\textbf{Accuracy}}=1$.

For the speed-up criteria, we calibrate the score using the $log_{10}$ function by using an adequate threshold of maximum speed-up to be reached for the task, meaning
\begin{equation}
\label{eq: score_ML_speed}
Score_{\textbf{Speed}}=min\Biggl(\Biggl(\frac{log_{10}(SpeedUp)}{log_{10}(SpeedUpMax)}\Biggl),1\Biggl),
\end{equation}
where
\begin{itemize}
\item $SpeedUp$ is given by
\begin{equation}
\label{eq: speedUp}
SpeedUp=\frac{time_{PhysicalSolver}}{time_{Inference}},
\end{equation}
\item $SpeedUpMax$ is the maximal speed up allowed for the airfoil use case
\item $time_{ClassicalSolver}$, the elapsed time to solve the physical problem using the classical solver
\item $time_{Inference}$, the inference time.
\end{itemize}

In particular, there is no advantage in providing a solution whose speed exceeds $SpeedUpMax$, as one would get the same perfect score $Score_{\textbf{Speed}}=1$ for a solution such that $SpeedUp=SpeedUpMax$.

Note that, while only the inference time appears explicitly in the score computation, it does not mean the training time is of no concern to us. In particular, if the training time overcomes a given threshold (for instance 72 hours on a single GPU), the proposed solution will be rejected. Thus, it would be equivalent to a null global score.

\subsection{Physical compliance score calculation}
While the machine learning metrics are relatively standard, the physical metrics are closely related to the underlying use case and physical problem. 
There are two physical quantities considered in this challenge namely: the drag coefficient and lift coefficient.
For each of them, we compute between the observations and predictions two coefficients: 
\begin{itemize}
    \item  The spearman correlation, a nonparametric measure of the monotonicity of the relationship between two datasets : 
    \begin{itemize}
    \item \verb|Spearman_correlation_drag| : $\rho_{D}$
    \item \verb|Spearman_correlation_lift| : $\rho_{L}$
    \end{itemize}
    \item The mean relative error:
    \begin{itemize}
    \item  \verb|mean_relative_drag| : $C_D$
    \item  \verb|mean_relative_lift| : $C_L$
    \end{itemize}
\end{itemize}

For the Physics compliance sub-score, we evaluate the  relative errors of physical  variables. For each criteria, the score is also calibrated based on 2 thresholds and gives 0 /1 / 2 points, similarly to $score_{\textbf{Accuracy}}$, depending on the result provided by the metric considered. 

\subsection{OOD generalization score calculation}
This sub-score will evaluate the capability on the learned model to predict OOD dataset. In the OOD testset, the input data are from a different distribution than those used for training. the computation of this sub-score is similar to $score_{ML}$ and is also based on two sub-criteria: accuracy and speed-up. To compute accuracy we consider the criteria used to compute the accuracy in $score_{ML}$ in addition to those  considered in physical compliance.

\subsection{Practical example}
Using the notation introduced in the previous subsection, let us consider the following configuration:
\begin{itemize}
\item $\alpha_{ML}=0.4$
\item $\alpha_{OOD}=0.3$
\item $\alpha_{PH}=0.3$
\item $\alpha_{A}=0.75$
\item $\alpha_{S}=0.25$
\item $SpeedUpMax = 10000$
\end{itemize}

In order to illustrate even further how the score computation works, we provide in table \ref{tab:Benchmark_evaluation_circle} two examples for the airfoil task: OpenFOAM (physical solver) and a fully-connected neural network (FC).   

\subsubsection{OpenFOAM score calculation}
As it is the most straightforward to compute, we start with the global score for the solution obtained with 'OpenFOAM', the physical solver used to produce the data . It is the reference physical solver, which implies that the accuracy is perfect but the speed-up is only equal to $1$ (no acceleration). Therefore, we obtain the following subscores
\begin{itemize}
\item $Score_{ML}= 0.75\times(\frac{2\times 5}{2\times 5})+ 0.25\times 0=0.75$
\item $Score_{OOD}=0.75\times(\frac{2\times 9}{2\times 9})+ 0.25\times 0=0.75$
\item $Score_{PH}=(\frac{2\times 4}{2\times 4})=1$
\end{itemize}

Then, by combining them, the global score is $Score_{OpenFOAM}=0.4\times 0.75+0.3\times 0.75+0.3\times 1=0.825$, therefore 82.5\%.

\subsubsection{FC score calculation}
The procedure is similar with 'FC' the associated subscores are:
\begin{itemize}
\item $Score_{ML}= 0.75\times(\frac{2\times 1 + 1}{2\times 5})+ 0.25\times\frac{log_{10}(750)}{log_{10}(10000)}\approx0.405$
\item $Score_{OOD}=0.75\times(\frac{2\times 1 + 1}{2\times 9})+ 0.25\times \times\frac{log_{10}(750)}{log_{10}(10000)}\approx0.305$
\item $Score_{PH}=(\frac{2\times 1}{2\times 4})=0.25$
\end{itemize}

For accuracy scores, the detailed results with their corresponding points are reported in table \ref{tab:detailed_results}. Speed-up scores are calculated using the equation \ref{eq: speedUp} as follows:
\begin{itemize}
\item $time_{PhysicalSolver}= 1500s$, $time_{Inference-ML} = 2s$ , $time_{Inference-OOD} = 2s$ 
\item $SpeedUp_{ML} = \frac{1500}{2} = 750$
\item $SpeedUp_{OOD} = \frac{1500}{2} = 750$
\end{itemize}
Then, by combining them, the global score is $Score_{FC}=0.4\times 0.405+0.3\times 0.305+0.3\times 0.25=0.3285$, therefore 32.85\%.


\begin{table}[t]
    \centering
    \vspace{-0.3cm}
    \caption{Scoring Table for the 3 tasks under 3 categories of evaluation criteria for the considered configuration. The performances are reported using three colors computed on the basis of two thresholds. Colors meaning:\begin{tabular}{ccc}\protect\scorecircle{100}{red} Unacceptable (0 point) & \protect\scorecircle{100}{orange} Acceptable (1 point)  &  \protect\scorecircle{100}{green}  Great (2 points) \end{tabular}.
    }
    \resizebox{\columnwidth}{!}{
    \begin{tabular}{clccccccccc}
    \toprule
         & & \multicolumn{9}{c}{\textbf{Criteria category}}\\
         & & \multicolumn{2}{c}{\textbf{ML-related (40\%)}}  && \multicolumn{2}{c}{\textbf{OOD generalization (30\%)}} && \multicolumn{1}{c}{\textbf{Physics (30\%)}} && \textbf{Score (100\%)}\\ \cline{3-4} \cline{6-7} \cline{9-9} \cline{11-11}
         & \textbf{Baseline} & Accuracy & Speed-up &&  OOD Accuracy & Speed-up  && Domain laws &&\\ \cline{1-11}
         \multicolumn{1}{c|}{\multirow{8}{*}{\rotatebox{90}{\texttt{\qquad\quad\ \ AirFoil}}}} & \multicolumn{1}{c|}{} & \underline{$\overline{u}_{x}$}\, \underline{$\overline{u}_{y}$}\, \underline{$\overline{p}$}\, \underline{$\overline{\nu}_{t}$}\, \underline{$\overline{p}_{s}$} & &&  \underline{$\overline{u}_{x}$} \underline{$\overline{u}_{y}$} \underline{$\overline{p}$} \underline{$\overline{\nu}_{t}$}  \underline{$\overline{p}_{s}$} \underline{$C_D$} \underline{$C_L$} \underline{$\rho_{D}$} \underline{$\rho_{L}$}  & && \underline{$C_D$} \underline{$C_L$} \underline{$\rho_{D}$} \underline{$\rho_{L}$} &&\\
         \multicolumn{1}{c|}{} & \multicolumn{1}{c|}{FC} & \myfullcircle{red}\myfullcircle{orange}\myfullcircle{red}\myfullcircle{green}\myfullcircle{red}  & 750  &&  \myfullcircle{red}\myfullcircle{orange}\myfullcircle{red}\myfullcircle{green}\myfullcircle{red} \myfullcircle{red}\myfullcircle{red}\myfullcircle{red}\myfullcircle{red}&  750 &&   \myfullcircle{red}\myfullcircle{orange}\myfullcircle{red}\myfullcircle{orange}&& 32.85\\ 
         \multicolumn{1}{c|}{} & \multicolumn{1}{c|}{OpenFOAM} & \myfullcircle{green}\myfullcircle{green}\myfullcircle{green}\myfullcircle{green}\myfullcircle{green} & 1  &&  \myfullcircle{green}\myfullcircle{green}\myfullcircle{green}\myfullcircle{green}\myfullcircle{green} \myfullcircle{green}\myfullcircle{green}\myfullcircle{green}\myfullcircle{green}&  1 &&   \myfullcircle{green}\myfullcircle{green}\myfullcircle{green}\myfullcircle{green}&& 82.5\\
         \bottomrule
    \end{tabular}}
    \label{tab:Benchmark_evaluation_circle}
    \vspace{-.5cm}

\end{table}

\section{Starting kit}
The provided starting kit \footnote{ \href{https://github.com/IRT-SystemX/ml4physim\_startingkit}{https://github.com/IRT-SystemX/ml4physim\_startingkit  }} includes a set of Jupyter notebooks helping the challenge participants to better understand the use case, the dataset and how to contribute to this competition.

\begin{itemize}
    \item \verb|0-Basic_Competition_Information|: This notebook contains general information concerning the competition organization, phases, deadlines and terms. The content is the same as the one shared in the competition Codabench page. 
    \item \verb|1-Airfoil_design_basic_simulation|: This notebook aims to familiarize the participants with the use case and to facilitate their comprehension. It allows the visualization of some simulation results. 
    \item \verb|2-Import_Airfoil_design_Dataset|: Shows how the challenge datasets could be downloaded and imported using proper functions. These data will be used in the following notebook to train and evaluate an augmented simulator. 
    \item \verb|3-Reproduce_baseline_results|: This notebook shows how the baseline results could be reproduced. It includes the whole pipeline of training, evaluation and score calculation of an augmented simulator using LIPS platform. 
    \item \verb|3b-Reproduce_baseline_results_Advanced_Configuration|: This notebook shows how another baseline results could be reproduced. It  also includes the whole pipeline of training, evaluation and score calculation of an augmented simulator using LIPS platform.
    \item \verb|4-How_to_Contribute|: This notebook shows 3 ways of contribution for beginner, intermediate and advanced users. The submissions should respect one of these forms to be valid and also to enable their proper evaluation through the LIPS platform which will be used for the final evaluation of the results. 
    
    \begin{itemize}
        \item Beginner Contributor: You only have to calibrate the parameters of existing augmented simulators
        \item Intermediate Contributor: You can implement an augmented simulator respecting a given template (provided by the LIPS platform)
        \item Advanced Contributor: you can implement your architecture independently from LIPS platform and use only the evaluation part of the framework to assess your model performance.
    \end{itemize}
    \item \verb|4a-How_to_Contribute_Tensorflow|: This notebook shows how to contribute using the existing augmented simulators based on Tensorflow library. The procedure to customize the architecture is fairly the same as pytorch (shown in Notebook 4).
    \item \verb|5-Scoring|: This notebook shows firstly how the score is computed by describing its different components. Next, it provides a script which can be used locally by the participants to obtain a score for their contributions. We encourage participants to evaluate their solutions via codabench (which uses the same scoring module as the one described in this notebook).
     \item \verb|6-Submission|: This notebook presents the composition of a submission bundle for Codabench and usable parameters.
     \item \verb|7-Submission_examples|: This notebook shows how to submit on Codabench and examples of submissions bundles.
    \\
    \\
\end{itemize}

\begin{table}[H]
    \centering
    \vspace{-0.9cm}
    \caption{Accuracy scores calculation of the FC solution.}
    \begin{tabular}{cccccc}
    \toprule
    Category & Criteria & obtained results & Thresholds & min/max & obtained score \\

     \cline{1-6} 
     \addlinespace
     \multirow{5}{*}{ML-Related} & $\overline{u}_{x}$\ & 0.208965 & T1=0.1 / T2 =0.2 & min & \myfullcircle{red} - 0 point \\
     & $\overline{u}_{y}$\ & 0.144508 &  T1=0.1 / T2=0.2 & min & \myfullcircle{orange} - 1 point \\
     & $\overline{p}$\ & 0.193066 & T1=0.02 / T2=0.1 & min & \myfullcircle{red} - 0 point \\ 
     & $\overline{\nu}_{t}$\ & 0.277285 & T1=0.5 / T2=1.0 & min & \myfullcircle{green} - 2 points  \\
     & $\overline{p}_{s}$ & 0.425576 & T1=0.08 / T2 =0.2  & min& \myfullcircle{red} - 0 point  \\
     \cline{1-6} 
          \addlinespace
      \multicolumn{6}{r}{$N$ = 5, $Nr$ = 3, $No$ = 1, $Ng$ = 1.   } \\
    \toprule 
    \toprule
         \addlinespace
     \multirow{9}{*}{OOD Generalization} & $\overline{u}_{x}$\ & 0.322766 & T1=0.1 / T2 =0.2 & min & \myfullcircle{red} - 0 point \\
     & $\overline{u}_{y}$\ & 0.199635 &  T1=0.1 / T2=0.2 & min & \myfullcircle{orange} - 1 point \\
     & $\overline{p}$\ & 0.333169 & T1=0.02 / T2=0.1 & min & \myfullcircle{red} - 0 point \\ 
     & $\overline{\nu}_{t}$\ & 0.431288 & T1=0.5 / T2=1.0 & min & \myfullcircle{green} - 2 points  \\
     & $\overline{p}_{s}$ & 0.805426 & T1=0.08 / T2 =0.2  & min& \myfullcircle{red} - 0 point  \\
     & $C_D$ & 21.793367 & T1=1 / T2 =10  & min& \myfullcircle{red} - 0 point  \\
     & $C_L$ & 0.711271 & T1=0.2 / T2 =0.5  & min& \myfullcircle{red} - 0 point  \\
     & $\rho_{D}$ & -0.043979 & T1=0.5 / T2 =0.8  & max& \myfullcircle{red} - 0 point  \\
     & $\rho_{L}$ & 0.917206 & T1=0.94 / T2 =0.98  & max& \myfullcircle{red} - 0 point  \\
     \cline{1-6} 
          \addlinespace
      \multicolumn{6}{r}{$N$ = 9, $Nr$ = 7, $No$ = 1, $Ng$ = 1.  } \\
        \toprule
        \toprule

             \addlinespace
     \multirow{4}{*}{Physical compliance } 
     
     & $C_D$ &16.345740 & T1=1 / T2 =10  & min& \myfullcircle{red} - 0 point  \\
     & $C_L$ & 0.365903 & T1=0.2 / T2 =0.5  & min& \myfullcircle{orange} - 1 point  \\
     & $\rho_{D}$ & -0.043079 & T1=0.5 / T2 =0.8  & max& \myfullcircle{red} - 0 point  \\
     & $\rho_{L}$ & 0.957070 & T1=0.94 / T2 =0.98  & max& \myfullcircle{orange} - 1 point  \\
     \cline{1-6} 
          \addlinespace
      \multicolumn{6}{r}{$N$ = 4, $Nr$ = 2, $No$ = 2, $Ng$ = 1.  } \\
       \toprule
        \toprule
    \end{tabular}
    \label{tab:detailed_results}
    \vspace{-.5cm}

\end{table}
 
\section{Conclusion}
In this paper, we introduce the ML4PhySim challenge, which aims to promote the use of ML-based surrogate models for solving physical problems. The competition focuses on a specific Computational Fluid Dynamics (CFD) use case: Airfoil design. It seeks to improve baseline solutions for the airfoil design use case by designing new ML-based surrogate models. The overarching goal is to optimize the trade-off between solution accuracy and computational cost, while also considering Out-of-Distribution (OOD) generalization and adherence to some basic physical constraints. The online training and evaluation of submitted solutions will establish a unified procedure for comparing and ranking entries. We hope that this competition will stimulate the development of novel machine learning approaches for tackling PDE-based physical problems and contribute to the long-term effort of developing benchmarks for real-world physical problems.

\newpage


\bibliographystyle{unsrt}  
\bibliography{references}

\end{document}